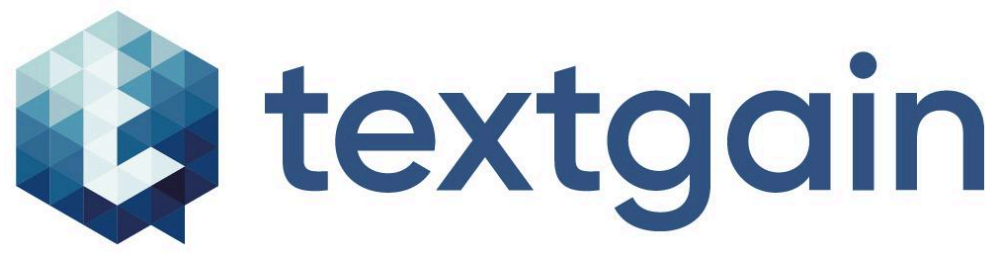

# Online antisemitism across platforms

Tom De Smedt, December 2021

**Abstract:** We created a fine-grained AI system for the detection of antisemitism. This Explainable AI will identify English and German antisemitic expressions of dehumanization, verbal aggression and conspiracies in online social media messages across platforms, to support high-level decision making.

# 1 Introduction

Online antisemitism is on the rise again, fuelled by worldwide COVID-19 conspiracy theories[1] and extremist actors seeking to exploit societal chaos for ideological purposes.[2] It is important to monitor this development,[3] but it is also challenging to sift through the ever-expanding social media data to identify relevant content.[4] We created a fine-grained lexicon (shown in Figure 1) and an AI system for detecting online antisemitism. This report provides a summary of the resource and its application.

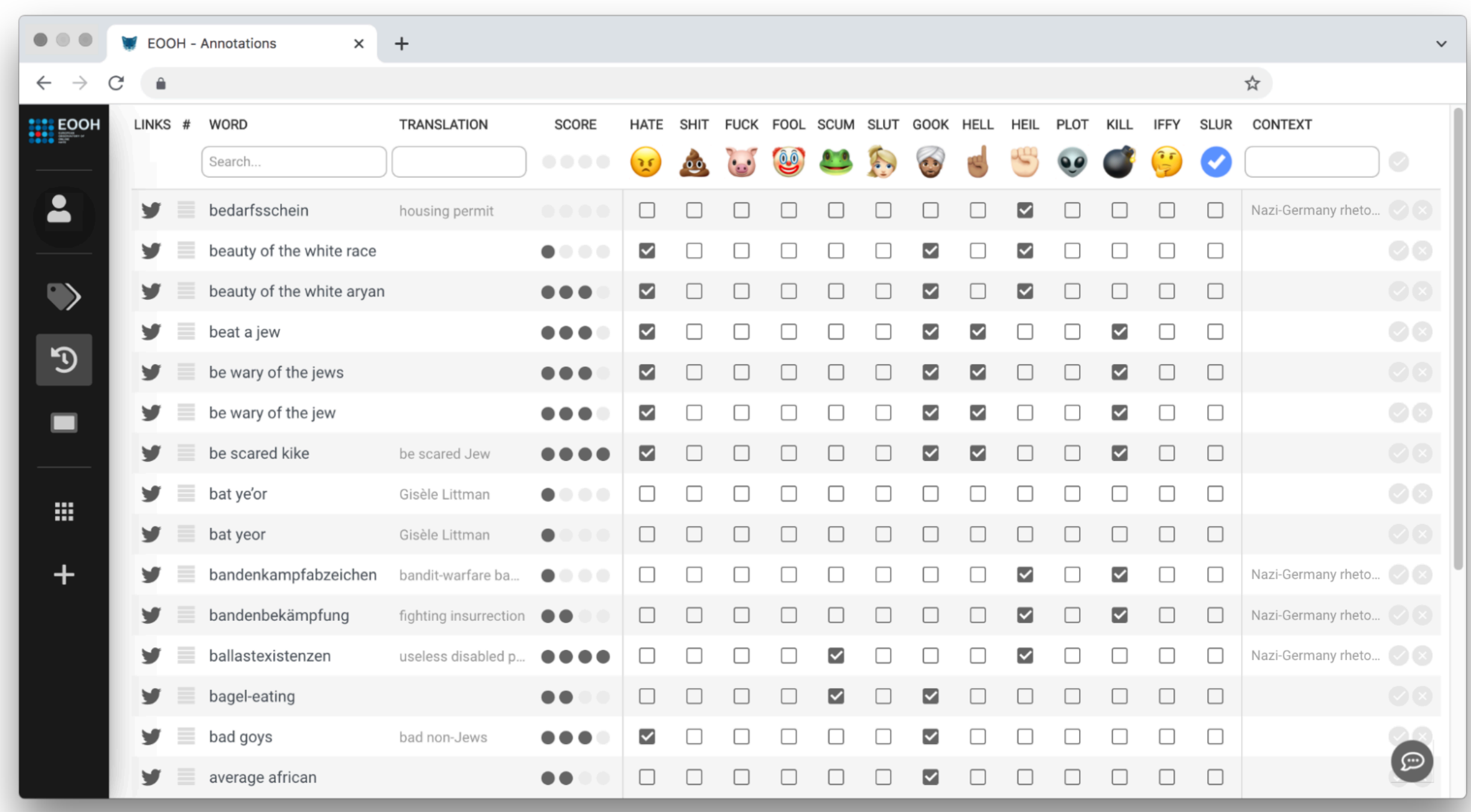

**Figure 1**. Fine-grained lexicon of antisemitic rhetoric, curated by multiple annotators.

# 2 Method & Materials

The lexicon contains over 2,000 relevant words and word combinations. This includes Nazi-Germany rhetoric (e.g., *Lügenpresse*, *Umvolkung*),[5] word combinations with dehumanising adjectives (e.g., *filthy* / *degenerate* + *Jew* / *Jewish*), word combinations with verbs inciting to violence (*gas* / *kill* the Jews), far-right terminology (e.g., *degenerate*, *subhuman*), observed alt-right neologisms (*holocough*, *lolocaust*, *Jewbacca*) and coded language (*14/88*, *HH*), and revived conspiracy theories (*blood libel*, *Thule Society*, *Vril*). Each entry is assigned a toxicity score between 0–4 by 2 annotators, where 0 means 'neutral' and 4 means 'extremely toxic'. Each entry can also have any combination of 10+ relevant labels, like RIDICULE, DEHUMANIZATION, VIOLENCE, RELIGION, POLITICS or CONSPIRACY (in our annotation tool represented by short 4-letter words like FOOL, SCUM, KILL, etc.).

---

[1] https://op.europa.eu/en/publication-detail/-/publication/d73c833f-c34c-11eb-a925-01aa75ed71a1

[2] https://op.europa.eu/en/publication-detail/-/publication/49e2ecf2-eae9-11eb-93a8-01aa75ed71a1

[3] https://fra.europa.eu/sites/default/files/fra_uploads/fra-2021-antisemitism-overview-2010-2020_en.pdf

[4] https://fra.europa.eu/sites/default/files/fra_uploads/fra-2018-experiences-and-perceptions-of-antisemitism-survey_en.pdf

[5] https://en.wikipedia.org/wiki/Glossary_of_Nazi_Germany

An AI system will then harness the lexicon to assign a toxicity score between 0–100 to social media messages, based on the words and word variations it contains,[6] indicating whether the message is likely to be antisemitic. The AI will also indicate *why* it thinks this is the case, by highlighting word combinations in the message. This is a form of Explainable AI (XAI) that is essential for high-stakes decision making.[7] Accompanying algorithms can anonymize messages in compliance with the GDPR across multiple social media platforms, making the AI system useful for real-time monitoring.

## 3 Results

To test the effectiveness of the AI system, we analysed 5 sets of social media messages on different platforms where we expect that antisemitic content is less or more likely to occur (Figure 2).

As a baseline, we analysed approximately 100K public posts on well-known social media platforms in 2015 (Twitter & Facebook). The year was intuitively chosen as a turning point in the spread of online hate and disinformation. The EU Code of Conduct on illegal hate speech (2016) was not in effect yet, and the Trump presidency (2017-2021) was not yet framing journalism as *fake news* or *lying press*,[8] terms reminiscent of Nazi-Germany's concept of *Lügenpresse*. The alt-right forum 8chan (2014), now understood as a principal source of antisemitic content,[9] was not yet that widely studied.

| SOURCE | **Twitter + FB** | **Twitter** | **Twitter** | **Gab** | **8chan** |
|---|---|---|---|---|---|
| YEAR | 2015 | 2020 | 2021 | 2020 | 2020 |
| SAMPLE | 100K | 100K | 100K | 10K | 100K |
| TOPIC | random | QAnon | QAnon | random | random |
| TOXICITY | 2/100 | 10/100 | 5/100 | 15/100 | 30/100 |
| ANTISEMITIC | 0.0% | 3.0% | 0.5% | 7.5% | 15.0% |
| VIOLENT | 0.0% | 1.5% | 0.5% | 5.0% | 8.5% |

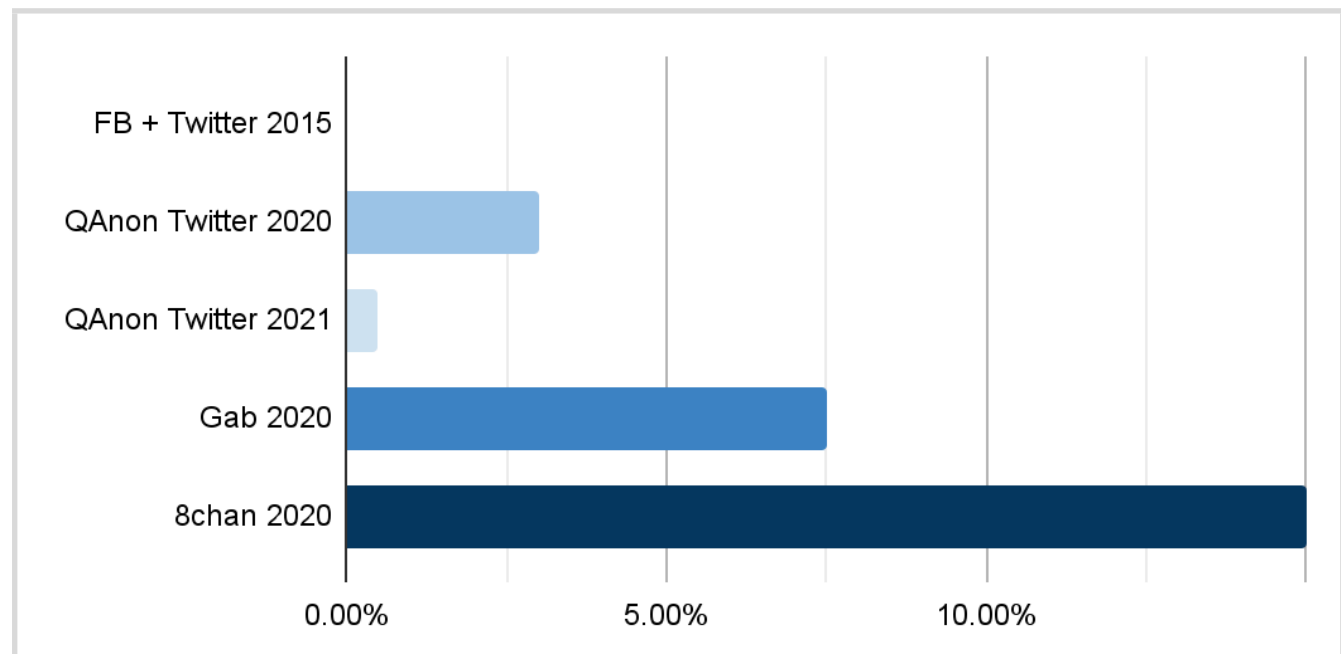

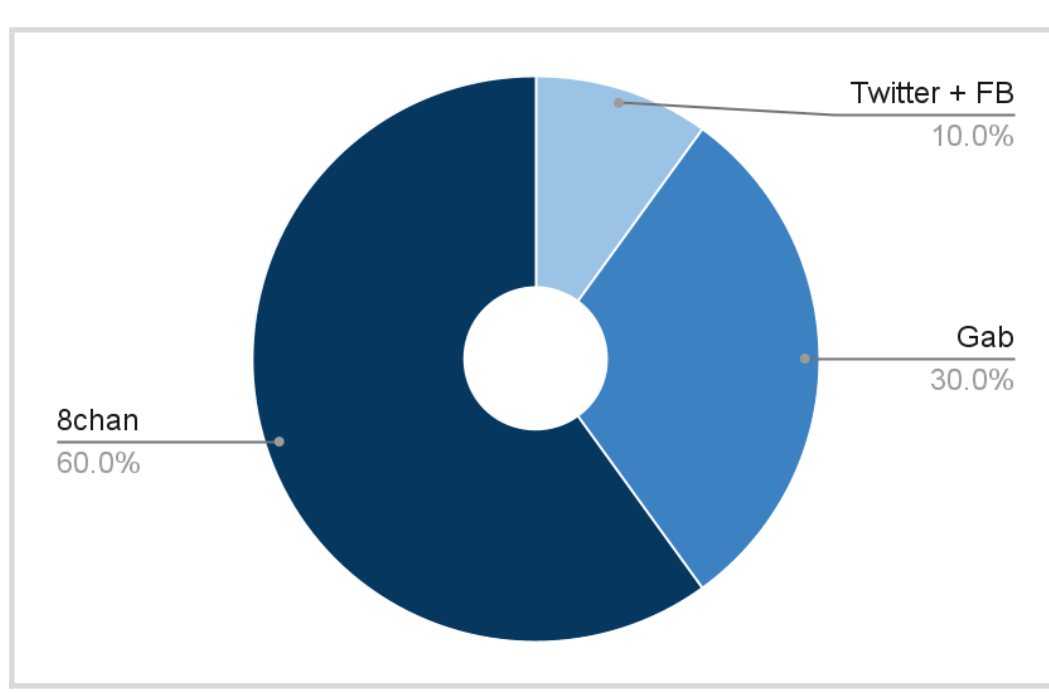

**Figure 2**. Percentage of antisemitic content across monitored platforms.

[6] De Smedt, T., Voué, P., Jaki, S., Röttcher, M., & De Pauw, G. (2020). Profanity & offensive words. *TGTR 3*.

[7] Rudin, C. (2019). Stop explaining black box machine learning models for high stakes decisions and use interpretable models instead. *Nature Machine Intelligence, 1*(5), 206-215.

[8] Quandt, T., Frischlich, L., Boberg, S., & Schatto-Eckrodt, T. (2019). Fake news. *The international encyclopedia of journalism studies*, 1-6.

[9] Voué, P., De Smedt, T., & De Pauw, G. (2020). 4chan & 8chan embeddings. arXiv:2005.06946.

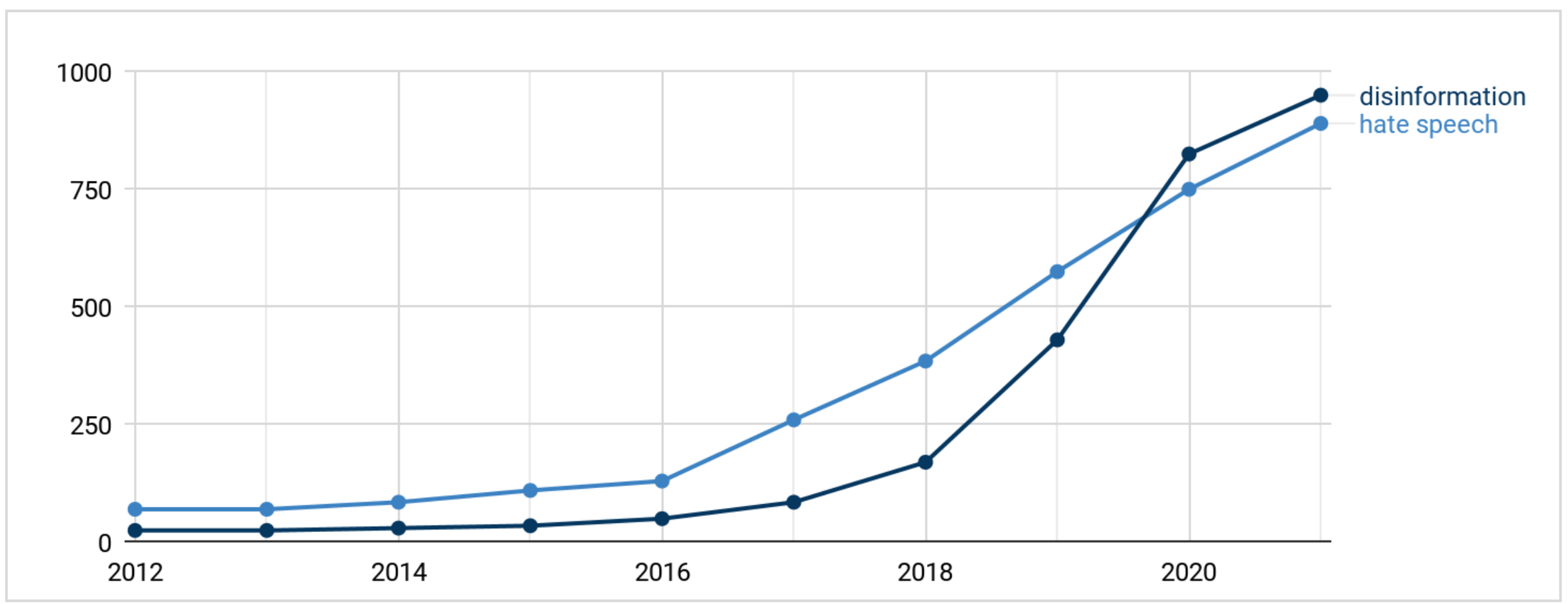

**Figure 3**. Academic publications on online hate speech and disinformation by year.

Conversely, 2015 saw the effectiveness of the Islamic State's online propaganda machine,[10] with its violent content fuelling far-right co-radicalisation targeting (Muslim) immigrants.[11] In 2016, *post-truth* was the Word of the Year.[12] In 2017, antisemitic QAnon beliefs spread from 8chan to mainstream social media platforms.[13] Jewish 'elites' are said to promote mass immigration in *Great Replacement* conspiracies, reviving the Nazi-Germany concept of *Umvolkung*.[14] In 2018, this culminated in the Pittsburgh synagogue shooting, for example.[15] By the end of 2019, QAnon conspiracies appear to mutate into a worldwide 'superconspiracy', propelled by a global pandemic and reviving antisemitic concepts such as *blood libel*.[16] In 2021, anti-government protests in Europe now have disgruntled citizens carrying QAnon flags, *Protocols of Zion* signs, or calling for a *Nuremberg 2.0 tribunal*.[17]

The number of scholarly publications on hate speech and disinformation then exponentially rises from dozens in 2015 to thousands in 2020 (Figure 3).[18]

**Twitter & Facebook, 2015**. Our baseline data consists of 100K random public posts written in English on Twitter and Facebook in 2015, collected using the Twitter API and the (pre-Cambridge Analytica) Facebook Graph API, where 'random' means posts with a punctuation mark like *?!.,* or any character like *aeiou*. The average toxicity of these posts is about 2/100, and we encountered less than 0.1% (1/1000) antisemitic posts. For the most part, these are 20 posts mentioning *Jews* and *kill them*.

**QAnon on Twitter, 2020**. Our 2020 data consists of 100K random QAnon posts written in English on Twitter, in the months before the Capitol riots (November–December 2020). The average toxicity is 10/100, and about 3% (1/30) posts are antisemitic, targeting *George Soros* and the *Rotschild*s.

---

[10] Lakomy, M. (2017). Cracks in the online "caliphate". *Perspectives on terrorism, 11*(3), 40-53.

[11] Pratt, D. (2016). Islam as feared other: perception and reaction. In *Fear of Muslims?* (pp. 31-43). Springer.

[12] Flood, A. (2016). "'Post-truth' named word of the year by Oxford Dictionaries'. The Guardian.

[13] Amarasingam, A., & Argentino, M. A. (2020). The QAnon conspiracy theory: A security threat in the making. *CTC Sentinel, 13*(7), 37-44.

[14] Dunst, C. ( 2018). "Is QAnon, the latest pro-Trump conspiracy theory, anti-Semitic?". Haaretz.

[15] Ingram, D. (2018). "Attacks on Jewish people rising on Instagram and Twitter, researchers say". NBC.

[16] https://www.researchgate.net/publication/348281072_The_QAnon_superconspiracy

[17] https://www.demorgen.be/politiek/en-toen-werd-antisemitisme-weer-heel-normaal-in-europa~b1aa5502

[18] https://app.dimensions.ai/discover/publication?search_mode=content&search_text=Hate%20speech&search_field=text_search

**QAnon on Twitter, 2021**. Our 2021 data consists of 100K random QAnon posts written in English on Twitter, in the months after the Capitol riots (January–February 2021). The average toxicity drops to 5/100, and about 0.5% (1/200) posts are antisemitic. We can give two possible reasons for this decrease. Firstly, Twitter suspended Trump's account along with 70,000 of the most prolific QAnon accounts.[19] This shows how removals might be effective to counter online hate and disinformation, alongside community self-regulation.[20] Secondly, QAnon followers seem to mask their language use after the riots, to avoid detection or legal repercussions, or they migrate to other platforms (Gab, Telegram).[21] This highlights the need to track phenomena across different platforms.

**Gab, 2020.** On Gab, we collected 10K random posts written in English in 2020. The average toxicity is 20/100, with 8.5% (1/12) antisemitic posts. The platform advertises itself as a free speech hub, but for a large part consists of far-right extremists and white supremacists banned from other platforms, and even the company itself has engaged in antisemitic remarks.[22] To illustrate this, the Pittsburgh synagogue shooting was publicly announced on Gab.[23] It is then unsurprising that the AI system ranks higher on its content (e.g., *Kalergi plan*, *kikes into soap*, *Jews rape kids*, see also Figure 4).

**8chan, 2020**. On 8chan (now 8kun), we collected 100K random messages written in English in 2020. This forum is anonymous and unmoderated: users post what they want without fear of repercussions or exposure to the real-life impact of their posts, and often attempt to outclass previous posts in viciousness.[24] The average toxicity is 30/100, with 15% (1/7) antisemitic messages. With a history of being a staging ground for cyberbullying and disinformation campaigns, and radicalisation to violent extremism, it begs the question why such forums continue to be online.

| KEYWORD | **Twitter+FB** | **QAnon 2020** | **QAnon 2021** | **Gab** | **8chan** |
|---|---|---|---|---|---|
| blood libel | ○○○ | ●●● | ○○○ | ○○○ | ○○○ |
| degenerates | ●○○ | ●○○ | ●○○ | ●●○ | ●●● |
| gas chamber | ○○○ | ○○○ | ●○○ | ●○○ | ○○○ |
| holohoax | ○○○ | ○○○ | ●○○ | ●●○ | ●●● |
| kikes | ○○○ | ○○○ | ●○○ | ●●● | ●●● |
| parasites | ○○○ | ●○○ | ●○○ | ●●○ | ●●○ |
| Rothschild | ○○○ | ●●○ | ●●○ | ●○○ | ●●○ |
| Soros | ○○○ | ●●○ | ●○○ | ●●○ | ●●○ |
| white genocide | ○○○ | ●○○ | ●○○ | ●●● | ●●● |

**Figure 4**. Example keywords and their prevalence (●○○ = dozens, ●●○ = hundreds, ●●● = thousands).

[19] https://www.bbc.com/news/technology-55638558

[20] De Smedt, T., Voué, P., Jaki, et al. (2021). Engagement analysis of counternarratives against online toxicity. arXiv:2111.07188.

[21] Hoseini, M., Melo, P., Benevenuto, F., Feldmann, A., & Zannettou, S. (2021). On the globalization of the QAnon conspiracy theory through Telegram. arXiv:2105.13020.

[22] Thalen, M. ( 2021). "Gab's CEO deactivates Twitter account after wildly antisemitic screed". The Daily Dot.

[23] Hutchinson, B., Levine, M., Weinstein, J., Seyler, M. (2018). "Alleged shooter posts on social media before attack". ABC News.

[24] Brown, A. (2018). What is so special about online (as compared to offline) hate speech?. *Ethnicities, 18*(3), 297-326.

## 4 Discussion

In summary, our AI system can correctly identify hubs with less or more antisemitic content. It can process thousands of messages in a matter of minutes, and it can be expanded easily with weak supervision to stay on top of evolving language use.

## 5 Acknowledgements

The development of the lexicon was co-funded by the European Commission and Get The Trolls Out. In 2019 the author was awarded the International Research Prize of the Auschwitz Foundation.

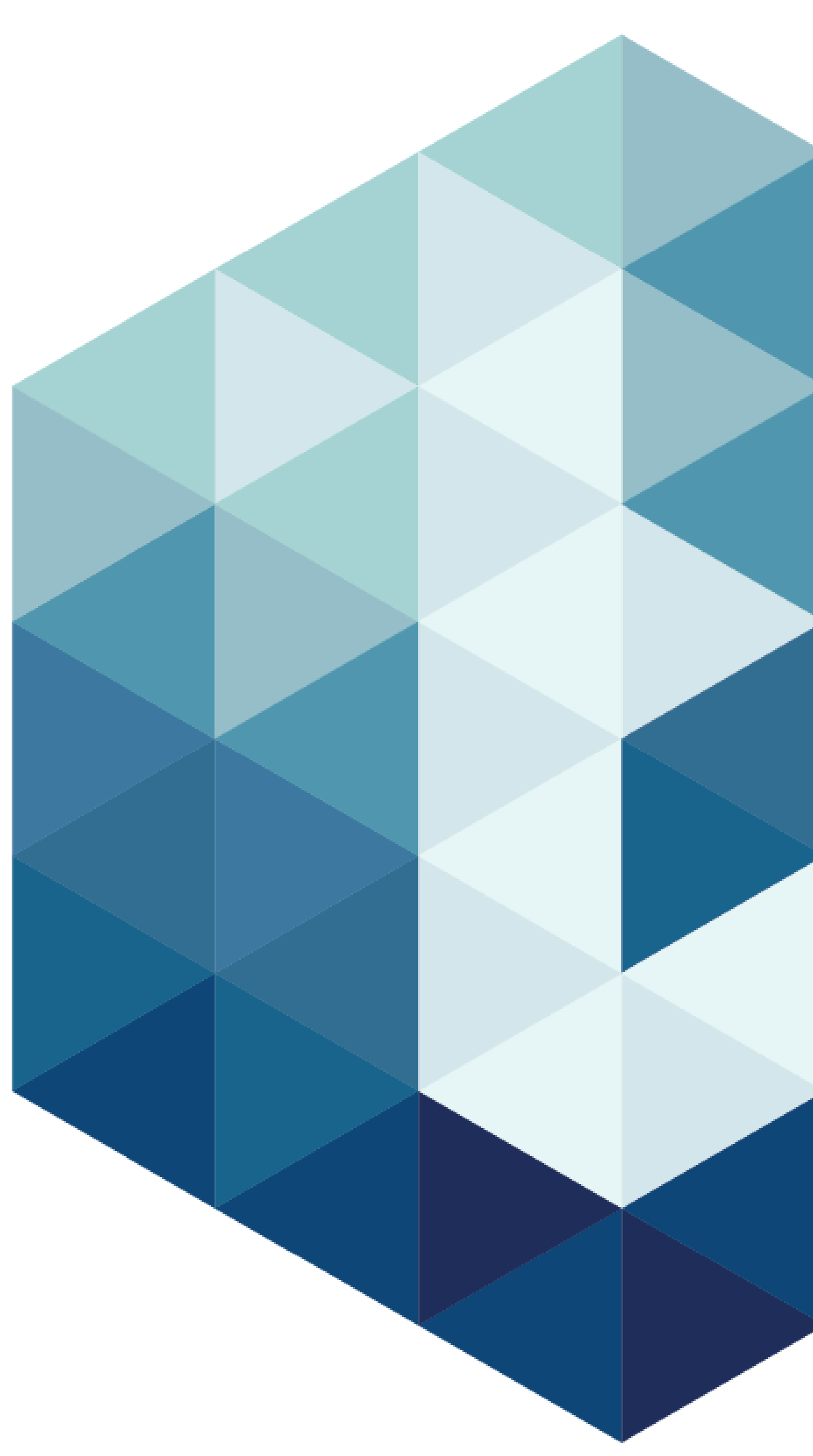

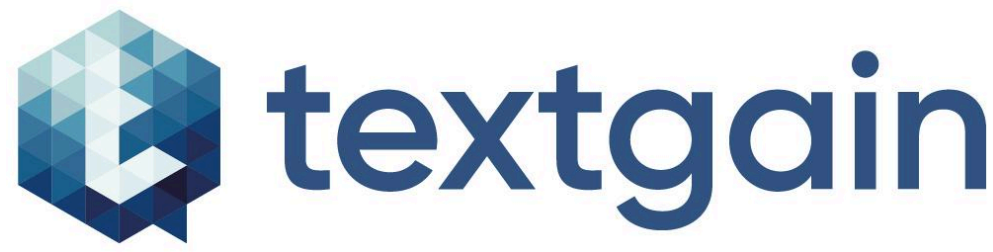

ARTIFICIAL INTELLIGENCE
THAT READS BETWEEN THE LINES

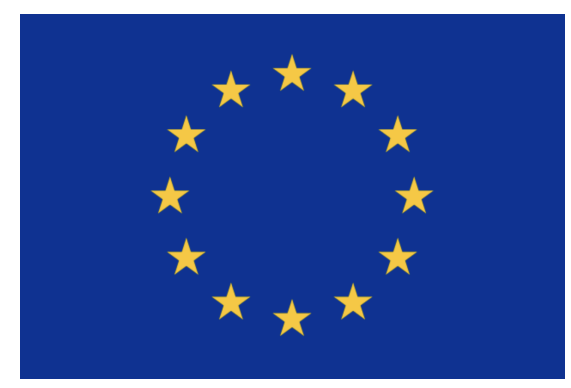